\documentclass[11pt]{article}

\usepackage[preprint]{acl}

\usepackage{times}
\usepackage{latexsym}
\usepackage{float}
\usepackage[T1]{fontenc}
\usepackage[utf8]{inputenc}
\usepackage{microtype}
\usepackage{inconsolata}
\usepackage{graphicx}
\usepackage{booktabs}
\usepackage{url}
\usepackage{graphicx}
\usepackage{subcaption}

\newcommand{\shortname}{\textit{VGBench}}

\newcommand{\QAbench}{\textit{VGQA}}
\newcommand{\Genbench}{\textit{VGen}}

\usepackage{graphicx} 
\title{\shortname{}: Evaluating Large Language Models on Vector Graphics Understanding and Generation}

\author{
  Bocheng Zou\thanks{Equal Contribution.} \quad Mu Cai$^{*}$  \quad Jianrui Zhang \quad Yong Jae Lee \\
  Department of Computer Science\\
  University of Wisconsin-Madison\\
  \texttt{\{bochengz,mucai,harrisz,yongjaelee\}@cs.wisc.edu} \\
  \vspace{-0.8em}\\
  \href{https://vgbench.github.io/}{https://vgbench.github.io/}
}

\begin{document}
\maketitle

\begin{abstract}

In the realm of vision models, the primary mode of representation is using pixels to rasterize the visual world. Yet this is not always the best or unique way to represent visual content, especially for designers and artists who depict the world using geometry primitives such as polygons. Vector graphics (VG), on the other hand, offer a textual representation of visual content, which can be more concise and powerful for content like cartoons, sketches and scientific figures. Recent studies have shown promising results on processing vector graphics with capable Large Language Models (LLMs). However, such works focus solely on qualitative results, understanding, or a specific type of vector graphics. We propose \shortname, a comprehensive benchmark for LLMs on handling vector graphics through diverse aspects, including (a) both visual understanding and generation, (b) evaluation of various vector graphics formats, (c) diverse question types, (d) wide range of prompting techniques, (e) under multiple LLMs and (f) comparison with VLMs on rasterized representations. Evaluating on our collected 4279 understanding and 5845 generation samples, we find that LLMs show strong capability on both aspects while exhibiting less desirable performance on low-level formats (SVG). Both data and evaluation pipeline will be open-sourced at \url{https://vgbench.github.io}.

\end{abstract}

\section{Introduction}

\begin{figure*}[h]
\centering
\includegraphics[width=\linewidth]{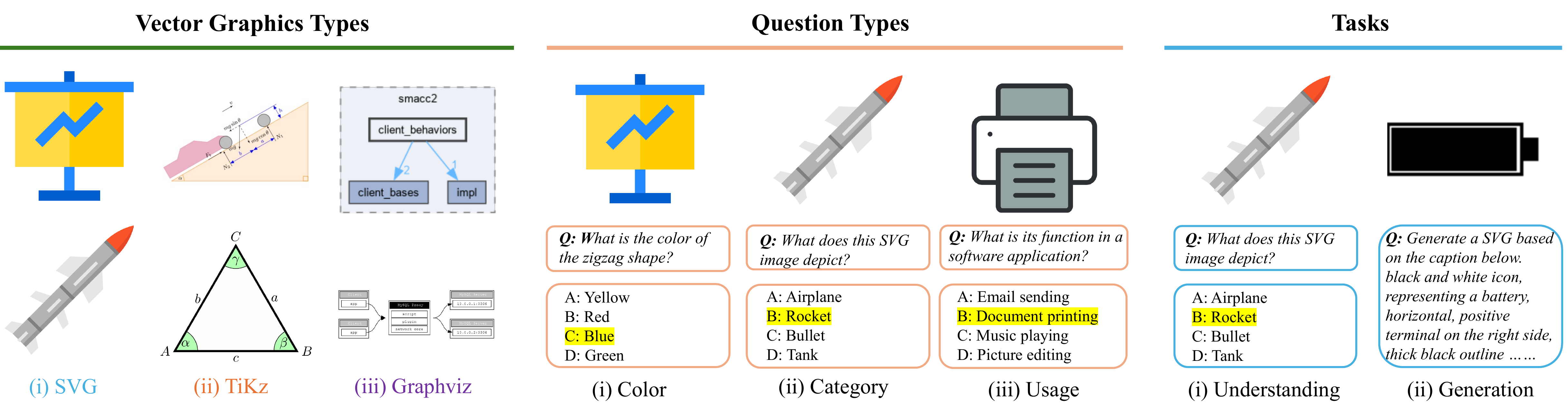}
\caption{\shortname{} is the \textbf{first comprehensive}  vector graphics (VG) \textit{understanding} and \textit{generation} benchmark across diverse vector graphics types, question types, and prompting techniques on a rich set of SoTA LLMs. Our large scale benchmark consists of 4279 multi-choice question-answer pairs and 5845 VG-caption pairs.}
\label{fig:tasks overview}
\end{figure*}

Current vision models are mostly built on pixels, rasterizing the visual world into a matrix representation. Such rasterized represents diverse visual content with equally sized elements. But pixels are not the only way to represent the visual world. For contents such as cartoons, sketches or scientific figures, a different representation using explicit geometry primitives can be more concise and beneficial. Vector graphics offer such a textual representation for visual content via geometry primitives, e.g., circles and polygons, as shown in Figure~\ref{fig:tasks overview} (a). Vector graphics have been critical for designers and artists since the geometry primitives can be easily manipulated. Vector representations include Scalable Vector Graphics~(SVG), TikZ, Graphviz, etc.

Vector Graphics vector representations make it possible to conduct visual understanding and generation with LLMs such as GPT-4~\cite{gpt4}. Recent studies~\cite{bubeck2023sparks, cai2023leveraging,rodriguez2023starvector} showcase LLMs' superior capability across different perspectives. However, those works either (1) only show qualitative results~\cite{bubeck2023sparks}, (2) only study vector graphics understanding~\cite{wang2024textbased} and not generation, or (3) only study one specific type of vector graphics such as SVG~\cite{cai2023leveraging, wang2024textbased, rodriguez2023starvector} or TikZ~\cite{belouadi2023auto}. Therefore, the community lacks a comprehensive LLM benchmark for vector graphics.

In this paper, we propose \shortname{} to comprehensively evaluate LLMs' vector graphics processing capabilities via different aspects: \shortname{} (1) includes both visual understanding (\QAbench{}) and generation (\Genbench{}); (2) evaluates diverse vector graphics formats such as SVG, TikZ, and Graphviz; (3) covers a set of taxonomies from low-level vision to high-level semantics, from color, shape, to category and advanced reasoning questions such as usage and the relation between objects; (4) adopts a variety of prompting techniques, such as zero-shot prediction, chain-of-thought reasoning, in-context learning, \textit{etc}.; (5) evaluates diverse LLMs including GPT-4~\cite{gpt4}, GPT-3.5~\cite{chatgpt}, Llama-3-8B-Instruct, Llama-3-70B-Instruct~\cite{llama-3}, Qwen2-7B-Instruct, Qwen2-72B-Instruct~\cite{qwen2}, Phi-3-mini-128k-instruct, Phi-3-medium-128k-instruct~\cite{abdin2024phi}, gemini-1.5-pro~\cite{reid2024gemini}; and (6) evaluates the VLM LLaVA-1.5-13b~\cite{liu2024improved} over rasterized representations of images in our benchmark..

We collect 4279 high-quality visual question-answer (QA) pairs for vector graphics (VG) understanding and 5845 VG-caption pairs for vector graphics generation. The vector graphics code is collected from existing datasets and the Internet. For visual question answering, we use a semi-automated pipeline to curate the questions. Specifically, we prompt GPT-4V(ision) to generate question-answer pairs given the provided in-context examples. Human annotators then filter the generated QA pairs to get the final high-quality vector graphics QA dataset. We use the gathered questions to evaluate if an LLM can understand vector graphics correctly. For text-to-vector-graphic generation (T2VG), we utilize GPT-4V to generate the captions and then use CLIP Score~\cite{hessel2021clipscore} and Fréchet Inception Distance (FID)~\cite{heusel2017gans} to evaluate the quality of the LLM generated vector graphics code. 

Our key findings are as follows:
\begin{itemize}
    \item LLMs show much better vector graphic understanding capability in TikZ and Graphviz than SVGs. TikZ and Graphviz include more high-level semantics compared to SVG, which is composed of low-level geometry primitives. This demonstrates that LLMs are more capable in understanding vector graphics code with high-level semantics.
    
    \item Advanced prompting techniques such as in-context learning or chain-of-thought prompting can bring significant performance boost for SVG, a low-level VG format. 

    \item LLMs show strong vector graphics generation ability on TikZ and Graphviz format compared to SVG format, hinting that TikZ or Graphviz might be a better medium for LLMs to manipulate vector graphics.
    
    \item In both understanding and generation, GPT-4 shows the strongest performance, yet open-source models such as Llama-3-70b shows competitive performance in understanding tasks. 
\end{itemize}

We hope that our work can serve as a foundation for LLM vector graphics understanding and generation benchmarking, and motivate further work to improve such capabilities. Our benchmark dataset and evaluation pipeline will be released.

\section{Related Work}

\subsection{Vector Graphics}
Vector graphics represent images using basic geometric elements like points, lines, and curves, rather than pixels. This method offers an alternative to raster graphics, providing advantages such as infinite scalability without losing detail and easy human manipulation.

There are a variety of vector graphics formats, such as SVG \cite{1218261}, TikZ \cite{mertz2007graphics} and Graphviz \cite{gansner2009drawing}. SVG format defines 14 functional areas or feature sets and represents graphics by recording basic information associated to these primitives, such as their coordination and scales, in an XML file. TikZ format defines some commands to build basic geometric elements and is mainly used with \LaTeX. In practice, third-party packages are also commonly used with TikZ to build more diverse images. Graphviz~\cite{gansner2009drawing} is a vector graphics format that focuses on representing different kinds of graphs. In this paper, we explore the said three kinds of vector graphics to provide a thorough and comprehensive analysis regarding the reasoning capabilities of LLMs on vector graphics.

\subsection{Evaluation for Image Understanding and Generation}

Works on Image Understanding are mainly based on raster images. VQA~\cite{antol2015vqa} first introduced the task of free-form and open-ended Visual Question Answering and evalauted existing LSTM-CNN based methods. CLIP~\cite{radford2021learning} introduces two encoders for both texts and images to achieve an aligned representation to serve as a baseline for many image understanding tasks. LLaVA \cite{liu2024visual} and LLaMA-Adapter~\cite{zhang2023llama} propose approaches to solve general-purpose visual and language understanding problems based on large language models.

While vector graphics can usually be converted to a raster image easily~\cite{gharachorloo1989characterization}, there are few works that try to directly understand the vector graphics format. \cite{jiang2021recognizing} explores such a way using graph neural networks. \cite{wang2024text} utilizes large language models to understand vector graphics. In our work, we utilize multiple prompting methods, to be mentioned in the following section, to evaluate different LLMs' vector graphics understanding capabilities by prompting them with the vector graphics code directly.

Most machine learning based image generation models aim to generate raster images~\cite{kingma2013auto, goodfellow2020generative, ho2020denoising, ramesh2021zero}. Some research focus on generating vector graphics in text format. Many works generate vector graphics from a raster image~\cite{diebel2008bayesian, xia2009patch, ha2017neural, ma2022towards}. Leveraging language models, some try to generate text representing vector graphics directly~\cite{carlier2020deepsvg, wu2023iconshop, rodriguez2023starvector}. In our work, we provide a different approach to evaluate vector graphics generation via leveraging competent multimodal models such as GPT4-V~\cite{gpt4} to generate a detailed caption from a rasterized image of a vector graphics object, based on which other LLMs will be generating vector graphics code for the same object during evaluation. We argue that models like GPT4-V can provide high-quality captions for us to automate part of the evaluation process.

\subsection{Prompting Techniques for Large Language Models}
\label{rel:prompt}
A variety of prompting strategies have been proven capable of boosting the performance of LLMs, such as GPT4~\cite{achiam2023gpt}. Few-shot learning~\cite{brown2020language} requires the user to give a few examples of the task to the LLM, while Chain of Thought~\cite{wei2022chain} instructs the LLM to think step by step to achieve higher performance. In-context learning~\cite{NEURIPS2020_1457c0d6} provides few-shot examples at inference time, and shows strong performance boost without updating the model's parameters. In this paper, we broadly evaluate LLMs' vector graphic understanding capability by employing the aforementioned prompting techniques.  

\begin{figure*}[t]
\centering
\includegraphics[width=\linewidth]{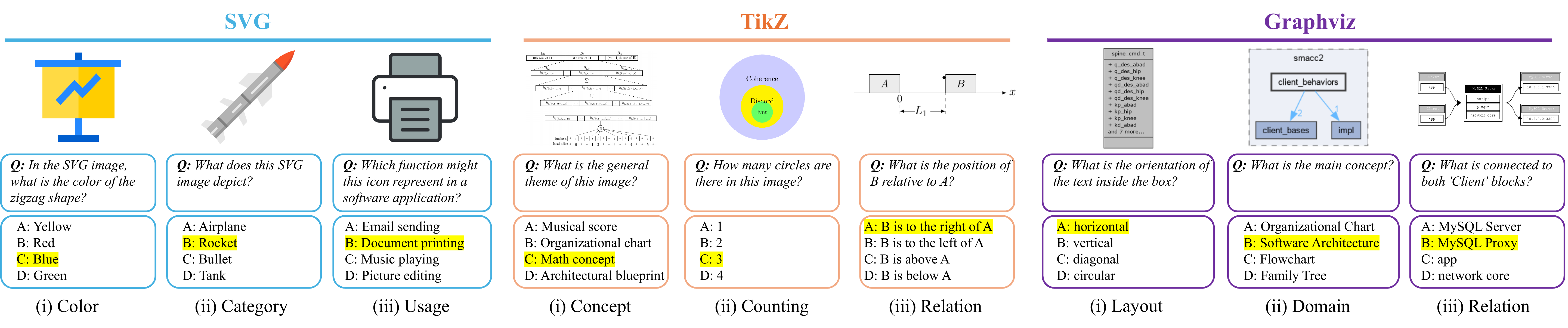}
\caption{Examples of the vector graphics QAs for diverse formats including SVG, TikZ, and Graphviz in \QAbench{}.}
\label{fig:QAtask example}
\end{figure*}

\begin{figure}[t]
\centering
\includegraphics[width=\linewidth]{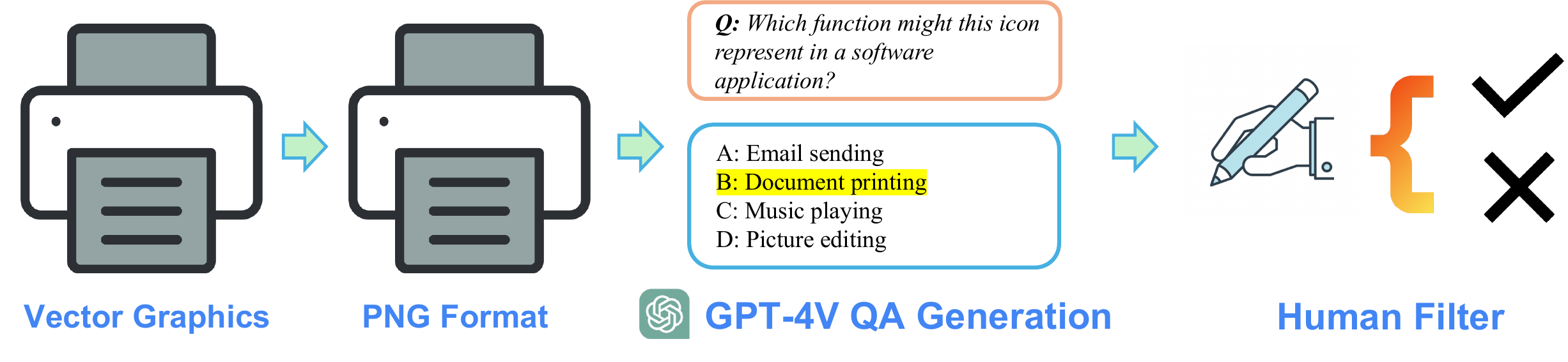}
\caption{The semi-automatic curation pipeline in \QAbench{}. Vector graphics are converted into PNG format, then GPT-4V is utilized to generate the questions and answers (QA) candidates. Finally, human annotators filter the QA pairs to obtain the high-quality QA dataset.  %
}
\label{fig:QA data curation}
\end{figure}

\begin{table*}[ht]
\centering
\resizebox{\textwidth}{!}{
\begin{tabular}{ccc|c||ccc|c||ccc|c}
\toprule
  \multicolumn{4}{c||}{\color{cyan}SVG} & \multicolumn{4}{c||}{\color{orange}TikZ} & \multicolumn{4}{c}{\color{violet}Graphviz} \\
\midrule
 \color{cyan}Category  & \color{cyan}Color & \color{cyan}Usage & Overall & \color{orange}Concept & \color{orange}Counting & \color{orange}Relation & Overall & \color{violet}Layout & \color{violet}Domain & \color{violet}Relation & Overall \\
 \midrule 869  & 671 & 688 & 2228 & 580 & 239 & 320 & 1139 & 319  & 418 & 175 & 912 \\
\bottomrule
\end{tabular}
}
\caption{Statistics of \QAbench{}. We collect a large set of QAs for each  vector graphics format under diverse tasks, resulting in 4279 QAs in total.  }

\label{tab:QA statistics}
\end{table*}

\begin{figure*}[h]
    \centering
    \captionsetup[subfigure]{labelfont={color=cyan},textfont={color=cyan}}
    \begin{subfigure}[b]{0.33\linewidth}
        \centering
        \includegraphics[trim=35 35 35 35,clip,width=\linewidth]{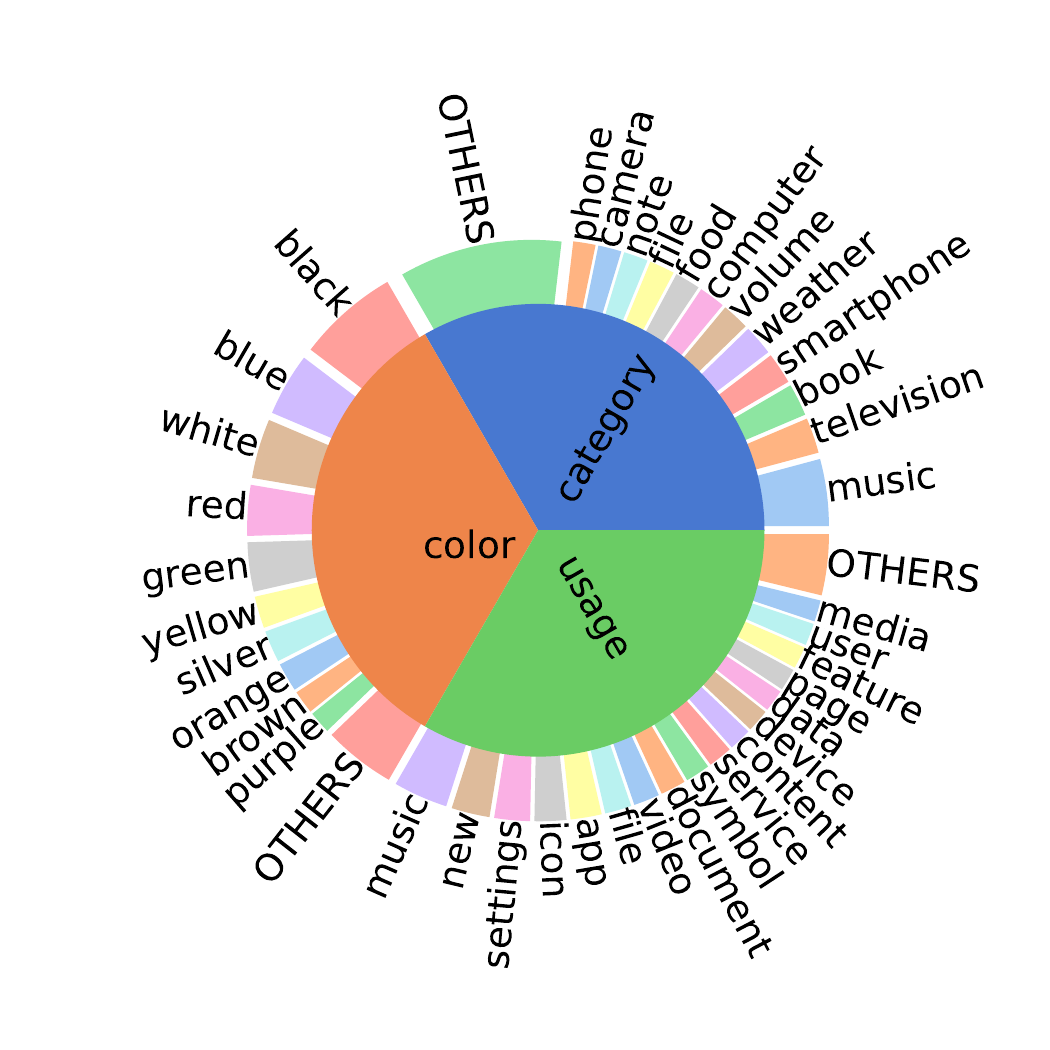}
        \caption{SVG}
    \end{subfigure}
    \captionsetup[subfigure]{labelfont={color=orange},textfont={color=orange}}
    \begin{subfigure}[b]{0.329\linewidth}
        \centering
        \includegraphics[trim=35 35 35 35,clip,width=\linewidth]{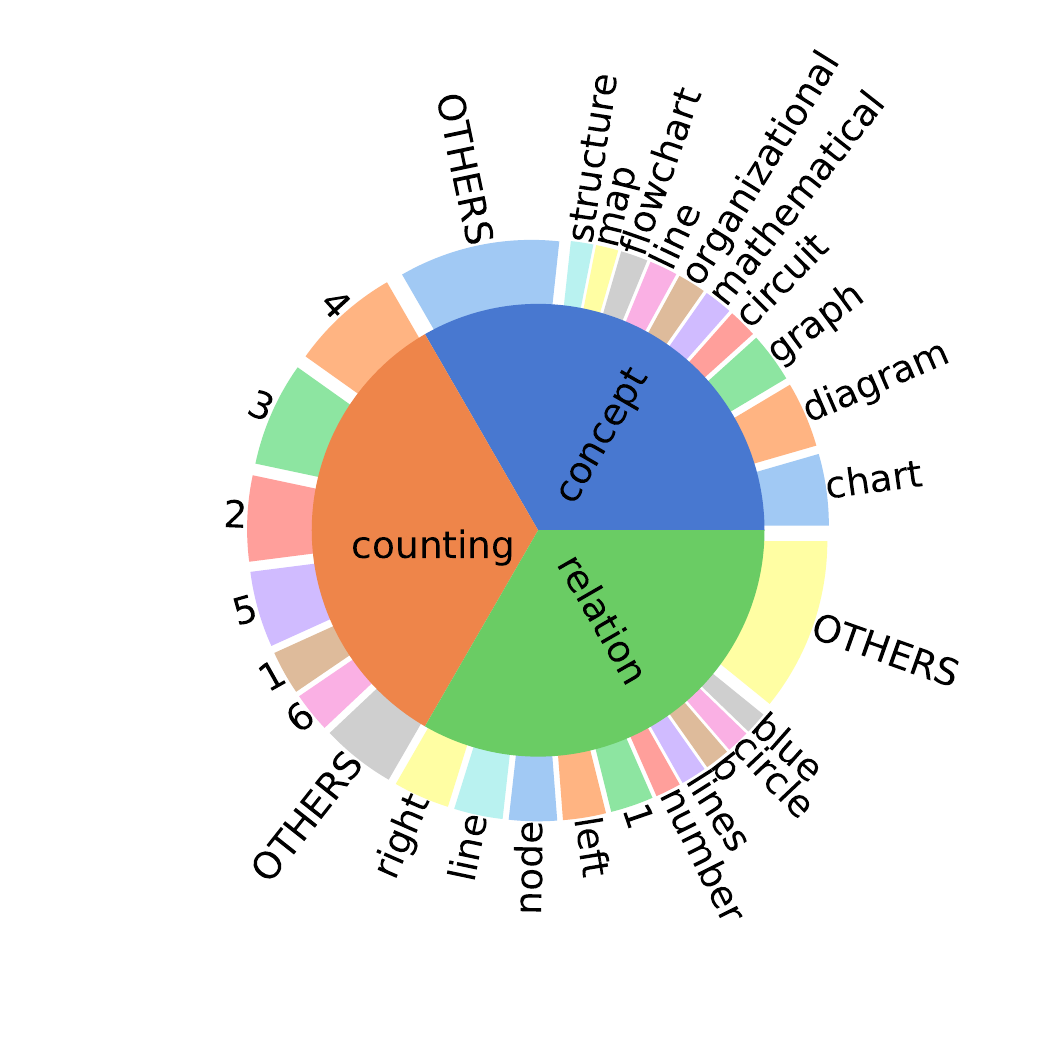}
        \caption{TikZ}
    \end{subfigure}
    \captionsetup[subfigure]{labelfont={color=violet},textfont={color=violet}}
    \begin{subfigure}[b]{0.33\linewidth}
        \centering
        \includegraphics[trim=35 35 35 35,clip,width=\linewidth]{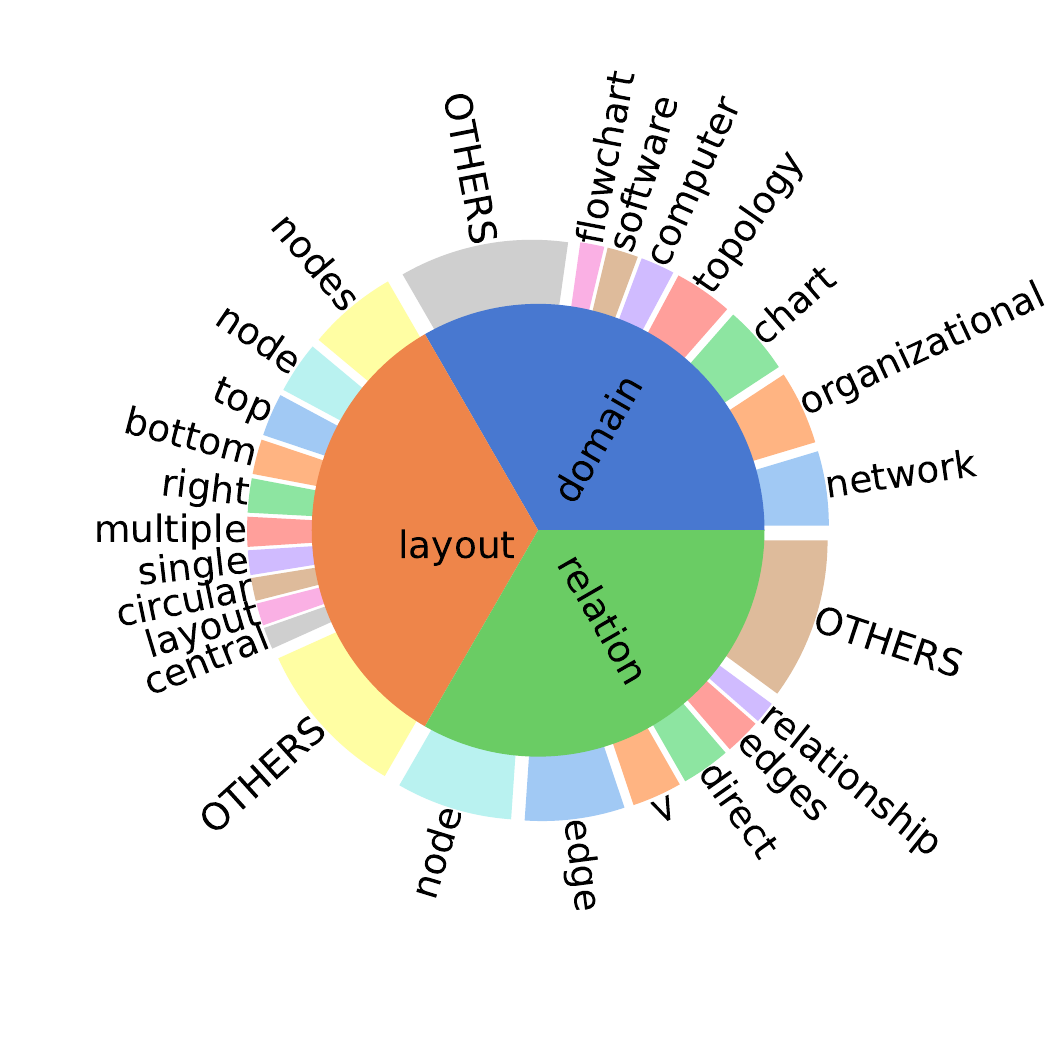}
        \caption{Graphviz}
    \end{subfigure}
    \caption{Word distribution based on question categories for each vector graphic type. The top 20 words are sampled from the answers to each type of question. Words with a frequency of less than 4\% are represented as "OTHERS".}
    \label{fig:word_distribution}
\end{figure*}

\begin{table*}[ht]
\centering
\resizebox{\textwidth}{!}{
\begin{tabular}{ccc|c||ccc|c||ccc|c}
\toprule
  \multicolumn{4}{c||}{\color{cyan}SVG} & \multicolumn{4}{c||}{\color{orange}TikZ} & \multicolumn{4}{c}{\color{violet}Graphviz} \\
\midrule
  \color{cyan}Category & \color{cyan} Color & \color{cyan}Usage & Average & \color{orange}Concept & \color{orange}Counting & \color{orange}Relation & Average & \color{violet}Layout & \color{violet}Domain & \color{violet}Relation & Average \\
 \midrule
 86.9  & 67.1 & 68.8 & 74.3 & 58.0 & 47.8 & 32.0 & 45.6 & 58.0 & 83.6 & 31.8 & 57.0 \\

\bottomrule
\end{tabular}
}
\caption{Human filtering passing rate  of \QAbench{}. TikZ and Graphviz show lower and less than half passing rate than SVG, indicating that even SoTA models exhibit poor vector graphic understanding capabilities in certain areas.}
\label{tab:QA statistics human passing rate}
\end{table*}

\begin{table*}[ht]
\centering
\resizebox{\textwidth}{!}{%
\begin{tabular}{@{}c||ccc|c||ccc|c||ccc|c@{}}
\toprule
\multicolumn{1}{c||}{} & \multicolumn{4}{c||}{\color{cyan}SVG} & \multicolumn{4}{c||}{\color{orange}TikZ} & \multicolumn{4}{c}{\color{violet}Graphviz} \\
\cmidrule(l){2-5} \cmidrule(l){6-9} \cmidrule(l){10-13} 
Prompting & \color{cyan}Category & \color{cyan}Color & \color{cyan}Usage & Avg & \color{orange}Concept & \color{orange}Counting & \color{orange}Relation & Avg & \color{violet}Domain & \color{violet}Layout & \color{violet}Relation & Avg \\ \midrule
Zero-Shot & 41.2 & 72.8 & 50.6 & 54.9 & 89.4 & 77.5 & 76.0 & 81.0 & 84.6 & 82.3 & 86.6 & 84.5 \\
In-Context Learning & 49.4 & 74.1 & 61.4 & 61.6 & 89.4 & 75.0 & 77.0 & 80.5 & 86.0 & 82.3 & 87.8 & 85.4 \\
Chain of Thought & 49.2 & 77.5 & 53.4 & 60.0 & 89.3 & 77.3 & 78.4 & 81.7 & 86.3 & 82.9 & 86.0 & 85.1 \\ 
\bottomrule
\end{tabular}%
}
\caption{Evaluation of \QAbench{} across diverse vector graphics formats for GPT-4. It can be seen that GPT-4 performs better on higher-level semantics with TikZ and Graphviz than on lower-level SVGs. It can also be seen that using specific prompting techniques improves performance, especially with Chain of Thought prompting. %
}
\label{tab:QABench results all}
\end{table*}

\section{Tasks and Experiments}
We first introduce the source of our vector graphics images in Sec.~\ref{exp:Data Collection}, and then describe the experiment settings in Sec.~\ref{exp:Experiment Settings}. After that, we detail our tasks, benchmark creation, evaluation pipeline and results for vector graphics understanding and generation in Sec.~\ref{exp:QAbench} and Sec.~\ref{exp:Genbench}, respectively. 
Finally, we provide in-depth analyses on the performance under different LLMs, %
different sequence lengths, and reasoning processes in Sec.~\ref{exp:In Depth Analysis}. 

\subsection{Vector Graphics Data Collection}
\label{exp:Data Collection}

We collect vector graphics samples for both understanding tasks and generation tasks from a variety of sources. For samples in SVG format, we collect them from a large-scale SVG repository.\footnote{\url{https://www.kaggle.com/datasets/victorcondino/svgicons}} We sample the TikZ format vector graphics code from the DaTikZ dataset~\cite{belouadi2023auto}. We sample the Graphviz code used to build our dataset by crawling GitHub.\footnote{\url{https://github.com/}}

\subsection{Experiment Settings}
\label{exp:Experiment Settings}

\paragraph{Vector Graphics Types}
Here we consider three major types of vector graphics: Scalable Vector Graphics~(SVG), TikZ, and Graphviz. SVG is exceptionally versatile and suitable for web applications, allowing for detailed graphical representations that scale infinitely without loss of quality. This enables SVGs to theoretically represent any visual content including complex animations and interactive elements. TikZ, in contrast, is specifically tailored for creating high-precision scientific illustrations within LaTeX documents, offering a comprehensive suite of tools for detailed diagrammatic representations; it encompasses a broad spectrum of high-level semantics such as "circuit diagrams, complex mathematical illustrations, and structured diagrams". Graphviz, on the other hand, belongs to the family of automated graph drawing tools, which are optimized for generating diagrams from abstract descriptions and data structures, making it ideal for visualizing hierarchical information, such as state machines, organizational charts, and network infrastructures.

\begin{figure}[ht]
\centering
\includegraphics[width=\linewidth]{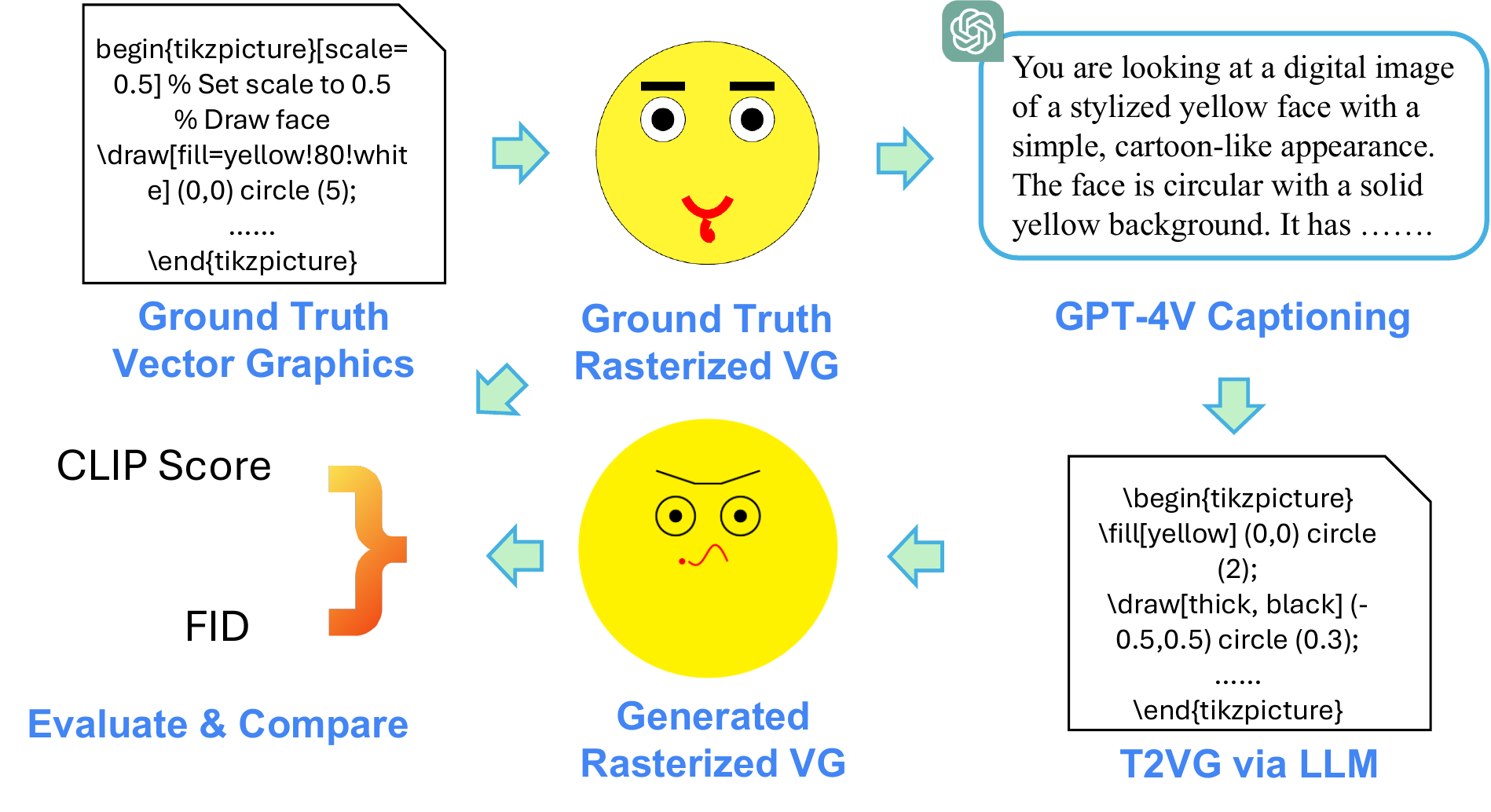}
\caption{The automatic generation pipeline in \Genbench{}. The vector graphics collected from the Internet is first rendered into the ground truth image then captioned by GPT-4V. The caption is fed into the target LLM to generate new vector graphics, which will be compared with the caption using CLIP Score and FID for a similarity score. The score is then compared with the similarity score between the ground truth and the same caption as the upper bound. }
\label{fig:generation}
\end{figure}

\begin{figure*}[ht]
\centering
\includegraphics[width=\linewidth]{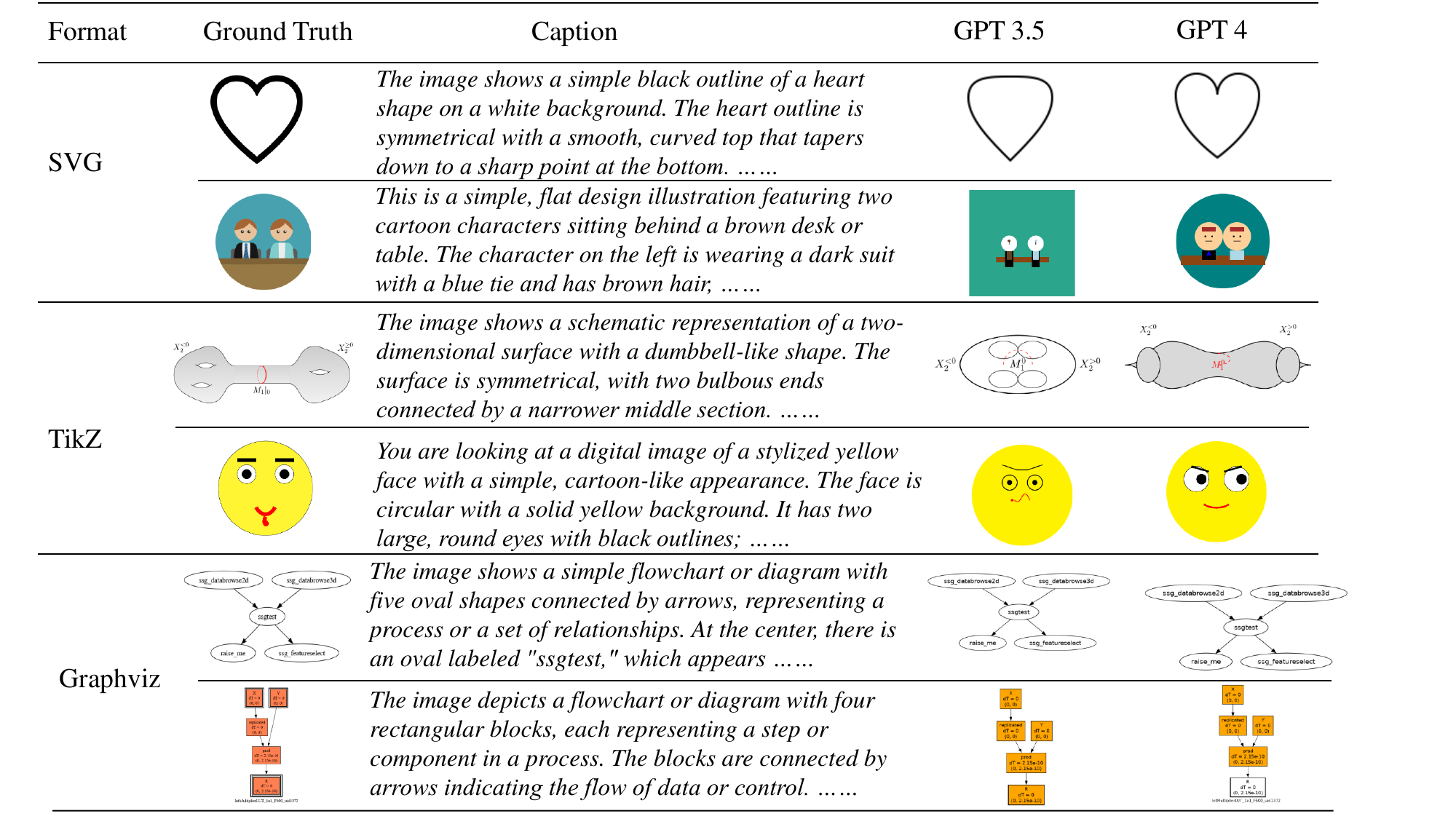}
\caption{Examples of generated vector graphics. The ground truth images are rendered by vector graphics code directly from Internet. The captions are generated by GPT-4V, The images on the right side are rendered by vector graphics code generated by GPT-3.5 or GPT-4.}
\label{fig:generation_examples}
\end{figure*}

\paragraph{Language Models}

We primarily use GPT-4 (1106 version)~\cite{gpt4} as the medium for vector graphics understanding and generation. This is because GPT-4 shows superior language reasoning and generation capabilities, as previously mentioned in Section~\ref{rel:prompt}. We also evaluated other proprietary models such as GPT-4o, GPT-3.5~\cite{chatgpt} and gemini-1.5-pro~\cite{reid2024gemini}, along with many other open-source LLMs that are highly capable, including Llama-3-70B-Instruct~\cite{llama-3}, Llama-3-8B-Instruct, Qwen2-7B-Instruct, Qwen2-72B-Instruct~\cite{qwen2}, Phi-3-mini-128k-instruct and Phi-3-medium-128k-instruct~\cite{abdin2024phi}.

\paragraph{Tasks} We consider two major tasks in computer vision: (1) visual understanding, and (2) visual generation. We design multiple choice questions to evaluate vector graphics understanding while using image generation metrics including Fréchet Inception Distance (FID)~\cite{heusel2017gans} and CLIP score~\cite{hessel2021clipscore} to measure the quality and correctness of generated vector graphics.

\paragraph{Prompting Techniques} We adopt three widely used prompting techniques: zero-shot, chain-of-thought (CoT) prompting, and in-context learning (few-shot prompting). For CoT, we instruct the LLM to think step by step by appending ``Please think step by step" to the initial question, using multi-round dialog to let the LLM consider each option separately before figuring out the answer. For in-context learning, we provide 3 examples of the same question type.

\subsection{\QAbench{}: Vector Graphics Understanding Benchmark}
\label{exp:QAbench}

\paragraph{Tasks}
\QAbench{} is designed to evaluate models' vector graphics understanding capability. We systematically design a range of tasks based on the nature of each vector graphics category, aiming at a comprehensive evaluation across different semantic levels. For SVG, we design three types of questions: color, category, and usage; for TikZ, we use concept, counting, and relations as types of questions; while for Graphviz, we design layout, domain, and relations. Examples are shown in Figure~\ref{fig:QAtask example}.

\paragraph{Benchmark Creation and Evaluation}
We employ a semi-automatic benchmark curation pipeline for \QAbench{}, as shown in Figure~\ref{fig:QA data curation}. Specifically, we render code representing vector graphics into a PNG image before leveraging GPT-4V~\cite{gpt4} to generate the 4-choice question-answer candidates. Then, human annotators with rich vision-linguistic knowledge make binary annotations to mark whether both the question and the answer of a candidate are rational, correct and belong to that specific type. Our approach brings several benefits: (i) annotation cost is greatly reduced due to GPT-4V's low API cost; (ii) GPT-4V is one of the most competitive LMMs that can provide high quality candidates; %
(iii) the human filtering process ensures the correctness of the final vector graphics understanding benchmark.

Finally, we collect 4279 samples in total, as shown in Table~\ref{tab:QA statistics} and~\ref{tab:QA statistics human passing rate}. The word distribution of answers in the \QAbench{} dataset is illustrated in Figure~\ref{fig:word_distribution}.  Specifically, we have 2228, 1139, and 912 samples for SVG, TikZ, and Graphviz, respectively.  After an LLM makes responses to the vector graphics questions, we compare the final responses with the ground-truth answers to compute accuracy. For LLMs with weaker instruction-following capabilities in producing easily parsable outputs, we use GPT-4 to determine their chosen option and then assess their accuracy.

\paragraph{Results}
Evaluation results of \QAbench{} under GPT-4~\cite{gpt4} are shown in Table~\ref{tab:QABench results all}. Several interesting findings arise from the results:

    \textbf{\textit{GPT-4 generally shows strong vector graphics understanding capability. }} In the zero-shot setting, GPT-4 shows non-trivial accuracy far beyond random accuracy (25\%) across all categories. Specifically, GPT-4 shows strong performance in TikZ, with an average accuracy of 78\%.
    
    \textbf{\textit{GPT-4 shows stronger performance in high-level vector graphics language (\textit{e.g.}, TikZ, Graphviz) compared to low-level vector graphics language SVG.} } In either zero-shot, few-shot, or Chain-of-Thought settings, TikZ and Graphviz show at least 17\% better performance than SVG. As a reminder, TikZ and Graphviz are fundamentally different from SVG in terms of the semantic levels, as SVG is composed of geometry primitives while TikZ and Graphviz contain high-level semantics such as ``above", ``below", explicit representation of nodes and edges, \textit{etc}. 
    
    \textbf{\textit{Chain of Thought (CoT) and In-Context Learning (ICL) show some performance improvements for some tasks, but not significant. } } CoT and ICL show $\sim$7\% performance boost for SVG which owns lowest performance among three formats. Yet CoT and ICL  show no benefits for TikZ and limited improvements for Graphviz, where GPT-4 already obtains $\sim$83\% accuracy under TikZ and Graphviz.
    
    \textbf{\textit{Different vector-graphic formats show diverse behaviors upon question types. } } For SVG, GPT-4 struggles at high-level questions and receives $\sim$50\% accuracy on category and reasoning types, while in TikZ and Graphviz, GPT-4 shows decent performance across all types of questions. This again demonstrates that GPT-4 shows inferior performance in low-level vector graphics tasks, especially on tasks related to reasoning.

\subsection{\Genbench{}: Vector Graphics Generation Benchmark}
\label{exp:Genbench}

\paragraph{Tasks}
We introduce \Genbench{}, a benchmark evaluating LLMs' vector-graphics generation capability. We use text to vector-graphics (T2VG) generation to test an LLM's ability to generate vector graphics code conditioned on a text prompt.

\paragraph{Benchmark Creation and Evaluation}
Again we evaluate on three vector graphics formats: SVG, TikZ and Graphviz. First, we obtain captions for each vector graphics image by leveraging GPT-4V~\cite{gpt4} over its rasterized image. Then we prompt the LLM to generate the vector graphics code corresponding to the caption.

\begin{table}[h]
\centering
\small
\begin{tabular}{l|ccc}
\toprule
             & \color{cyan}SVG                       & \color{orange}TikZ                      & \color{violet}Graphviz \\ \midrule
\# of VG-captions pairs & 2000 & 2000 & 1845 \\
\midrule
\end{tabular}
\caption{Statistics of \Genbench{} on three VG formats.}
\end{table}

Finally, we map the generated vector graphics into rasterzied images, then use CLIP Score and Fréchet Inception Distance (FID) Score to evaluate the quality of the generated vector graphics.

We use CLIP Score to measure the similarity between each generated vector graphics and its associated caption. We utilize Long-CLIP \cite{zhang2024long} instead of the vanilla CLIP~\cite{radford2021learning} since our detailed captions are often longer than CLIP's maximum context length of 77. FID is utilized to evaluate the distribution gap between the original vector graphics and generated ones. For both metrics, we use the score of our ground truths as the upper bound to reflect the quality of the generated images. The overall pipeline is shown in Figure~\ref{fig:generation}.

\begin{table}[h]
\centering
\small
\begin{tabular}{l|lll}
\toprule
      LLM        & \color{cyan}SVG                       & \color{orange}TikZ                      & \color{violet}Graphviz \\ \midrule
Ground-Truth & 25.61 & 24.63 & 23.67 \\
GPT-4        & 23.97 & 24.42 & 24.50 \\
GPT-3.5-Turbo & 22.88 & 24.21 & 23.88 \\ \bottomrule
\end{tabular}
\caption{CLIP score between captions and rasterized images from the generated vector graphics.  }
\label{clip_score}
\end{table}

\begin{table}[h]
\centering
\small
\begin{tabular}{l|lll}
\toprule
       LLM      & \color{cyan}SVG                       & \color{orange}TikZ                      & \color{violet}Graphviz \\ \midrule
GPT-4        & 44.81 & 39.38 & 77.03 \\
GPT-3.5-Turbo & 60.67 & 17.49 & 88.20 \\ \bottomrule
\end{tabular}
\caption{The FID score between the ground truth images and the generated images. Lower is better. }
\label{fid_score}
\end{table}

\begin{table*}[h]
\centering
\resizebox{\textwidth}{!}{%
\begin{tabular}{@{}c||ccc|c||ccc|c||ccc|c@{}}
\toprule
\multicolumn{1}{c||}{} & \multicolumn{4}{c||}{\color{cyan}SVG} & \multicolumn{4}{c||}{\color{orange}TikZ} & \multicolumn{4}{c}{\color{violet}Graphviz} \\
\cmidrule(l){2-5} \cmidrule(l){6-9} \cmidrule(l){10-13} 
Model & \color{cyan}Category & \color{cyan}Color & \color{cyan}Usage & Avg & \color{orange}Concept & \color{orange}Counting & \color{orange}Relation & Avg & \color{violet}Domain & \color{violet}Layout & \color{violet}Relation & Avg \\ \midrule
GPT-3.5-Turbo & 33.4 & 50.5 & 47.1 & 43.7 & 76.7 & 56.8 & 54.4 & 62.6 & 83.6 & 62.5 & 63.5 & 69.9 \\
GPT-4o & 52.5 & 80.4 & 60.3 & 64.4 & 87.0 & 75.0 & 77.3 & 79.8 & 83.6 & 75.0 & 83.7 & 80.8  \\
GPT-4 & 41.2 & 72.8 & 50.6 & 54.9 & 89.4 & 77.5 & 76.0 & 81.0 & 84.6 & 82.3 & 86.6 & 84.5 \\
Llama-3-8B & 32.3 & 39.8 & 48.0 & 40.0 & 64.6 & 53.0 & 45.9 & 54.5 & 68.0 & 52.5 & 55.8 & 58.8 \\ 
Llama-3-70B & 46.3 & 58.7 & 55.3 & 53.4 & 78.5 & 68.2 & 66.7 & 71.1 & 72.8 & 61.4 & 74.4 & 69.5 \\ 
Qwen2-7B & 33.3 & 48.7 & 46.3 & 42.8 & 79.4 & 64.7 & 58.3 & 67.5 & 81.8 & 57.3 & 68.6 & 69.2 \\ 
Qwen2-72B & 43.4 & 62.4 & 55.9 & 53.9 & 88.6 & 74.6 & 72.5 & 78.6 & 86.5 & 71.5 & 80.8 & 79.6 \\ 
Phi-3-Mini-128K & 34.1 & 29.8 & 49.7 & 37.9 & 70.6 & 52.5 & 50.7 & 57.9 & 74.7 & 58.9 & 68.6 & 67.4  \\ 
Phi-3-Medium-128k & 43.6 & 44.7 & 60.6 & 49.6 & 80.4 & 59.7 & 62.8 & 67.6 & 81.4 & 66.5 & 72.7 & 73.5   \\ 
Gemini-1.5-Pro & 39.2 & 73.2 & 47.9 & 53.4 & 86.7 & 74.9 & 71.8 & 77.8 & 79.5 & 66.8 & 86.0 & 77.4 \\ 
LLaVA-1.5-13b-hf & 83.0 & 85.2 & 84.0 & 84.1 & 64.3 & 34.3 & 44.8 & 47.8 & 46.7 & 53.9 & 49.7 & 50.1 \\

\bottomrule
\end{tabular}%
}
\caption{The evaluation of \QAbench{} across diverse vector graphics formats for different LLMs and the evaluation of rasterized representation of \QAbench{} in VLMs in the zero-shot setting. }
\label{tab:QABench results multi-llm}
\end{table*}

\begin{table*}[h]
\centering
\resizebox{\textwidth}{!}{%
\begin{tabular}{@{}c||ccc|c||ccc|c||ccc|c@{}}
\toprule
\multicolumn{1}{c||}{} & \multicolumn{4}{c||}{\color{cyan}SVG} & \multicolumn{4}{c||}{\color{orange}TikZ} & \multicolumn{4}{c}{\color{violet}Graphviz} \\
\cmidrule(l){2-5} \cmidrule(l){6-9} \cmidrule(l){10-13} 
Length & \color{cyan}Category & \color{cyan}Color & \color{cyan}Usage & Avg & \color{orange}Concept & \color{orange}Counting & \color{orange}Relation & Avg & \color{violet}Domain & \color{violet}Layout & \color{violet}Relation & Avg \\
\midrule
   1-1000 & 59.3 & 72.2 & 65.5 & 65.7 & 83.2 & 79.8 & 69.9 & 77.6 & 78.4 & 84.1 & 90.4 & 84.3 \\
1000-2000 & 47.0 & 75.3 & 60.8 & 61.0 & 88.0 & 83.8 & 79.7 & 83.8 & 90.4 & 77.6 & 87.9 & 85.3 \\
2000-3000 & 46.8 & 76.4 & 50.6 & 57.9 & 85.5 & 66.7 & 80.0 & 77.4 & 96.2 & 80.0 & 77.8 & 84.7 \\
3000-4000 & 51.5 & 64.1 & 54.1 & 56.6 & 89.7 & 69.2 & 70.0 & 76.3 & 90.5 & 82.4 & 75.0 & 82.6 \\ 
   > 4000 & 48.9 & 70.1 & 52.5 & 57.2 & 95.8 & 55.0 & 81.2 & 77.3 & 87.0 & 82.5 & 72.2 & 80.6\\ 
\bottomrule
\end{tabular}%
}
\caption{ \QAbench{}  performance under different lengths of vector graphics for GPT-4 with zero-shot prompting. GPT-4 performs better on some lengths than others. For instance, in the Graphviz Domain question type, GPT-4 performs at an outstanding 96\% accuracy on the 2k-3k range while showing most subpar performance on the <1k range.%
}
\label{tab:QABench length}
\end{table*}

\begin{figure*}[!ht]
\centering
\includegraphics[width=\linewidth]{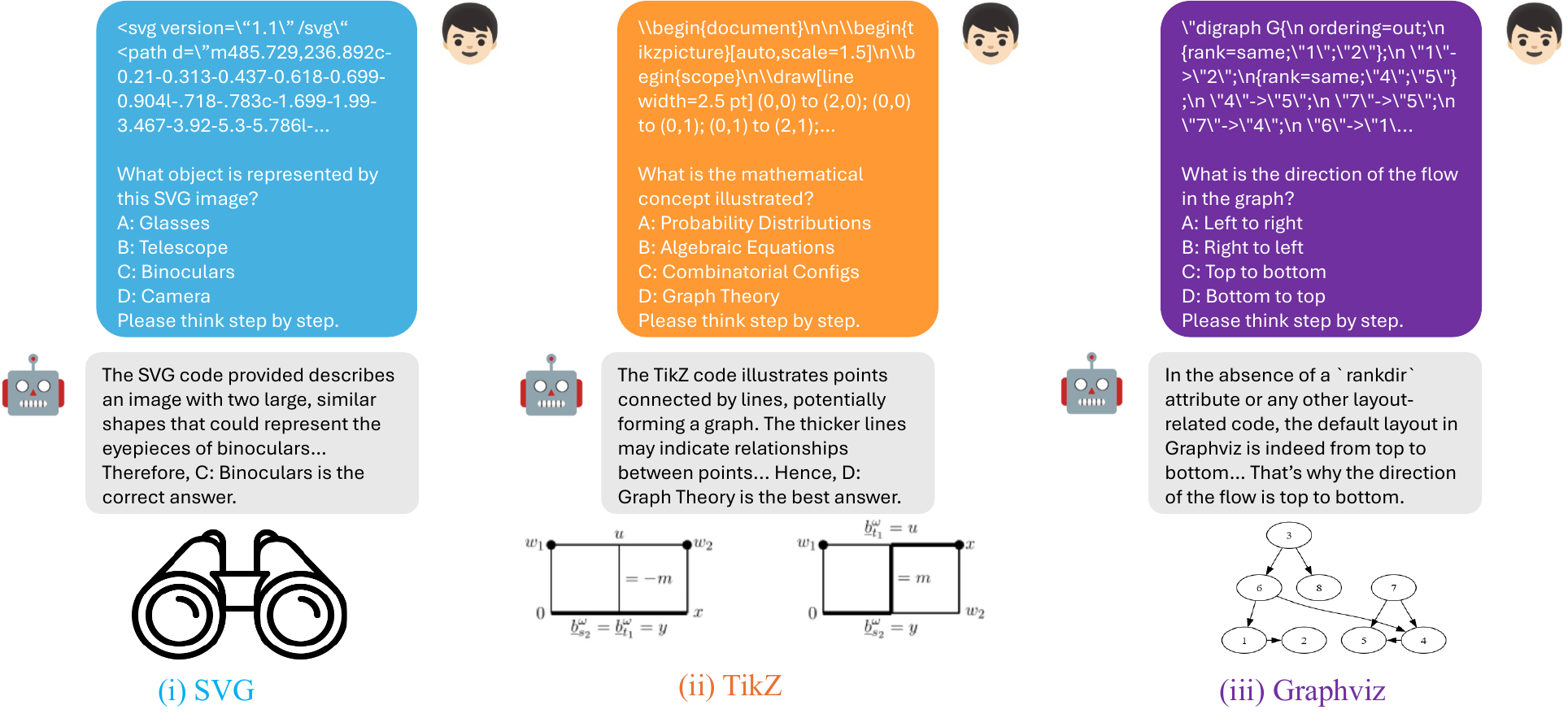}
\vspace{-2em}
\caption{Examples of prompting GPT-4 using Chain-of-Thought with different types of vector graphics in \QAbench{}. We only show GPT-4 the vector graphics code, but we include the rasterized images here for the sake of the readers.}
\label{fig:prompt example}
\end{figure*}

\paragraph{Results}

CLIP and FID score under each VG type is shown in Table~\ref{clip_score} and Table~\ref{fid_score}, respectively.  Qualitative examples are shown in Figure~\ref{fig:generation_examples}.

\textit{\textbf{Both GPT-3.5 and GPT-4 show strong vector graphics generation capability. }} Both LLMs show similar CLIP score as the ground truth. Results on FID score also support this claim. GPT-4 shows better performance than GPT-3.5 on CLIP score. Qualitative examples including the heart shape  and flowchart generation also demonstrate the promising capability of VG generation using LLMs.

\subsection{In Depth Analysis}
\label{exp:In Depth Analysis}

\paragraph{Impact of Different LLMs}

We next perform experiments over a variety of large language models, including GPT-4, GPT-3.5, Llama-3-70B-Instruct~\cite{llama-3} and Llama-3-8B-Instruct. Results are shown in Table~\ref{tab:QABench results multi-llm}. The results show that GPT-4 has the best VG understanding ability over vector graphics among those models while Llama-3-70B shows better performance than GPT-3.5. 

\paragraph{Comparison between LLMs and MLLMs on Image Understanding}

We find that VLMs such as LLaVA \cite{liu2023llava} show interesting behavior compared with LLMs on vector graphics. To evaluate the performance of VLMs on \QAbench{}, we render each visual content in our benchmark into the PNG format, and then feed the same question to VLMs. Specifically, we evaluate LLaVA-1.5-13b\cite{liu2024improved}, as shown in Table \ref{vlms}. LLaVA-1.5 shows strong performance in SVG format rather than TikZ and Graphviz. The strong performance gain that LLMs obtained on high-level vector graphics languages such as TikZ and Graphviz shows that those kinds of vector graphics are more aligned with LLMs' training data, natural languages, which is a highly compressed representation of the world. Low-level vector graphics languages such as SVG cover more low-level visual signals that can be better handled by VLMs using their rasterized representation.

\begin{table}[H]
\centering
\small
\begin{tabular}{l|lll}
\toprule
      LLM        & \color{cyan}SVG Avg & \color{orange}TikZ Avg& \color{violet}Graphviz Avg\\ \midrule
GPT-4& 54.9 & \textbf{81.0} & \textbf{84.5} \\
GPT-4o(Text)& 64.4 & 79.8 & 80.8 \\
LLaVA-1.5-13b& \textbf{84.1} & 47.8 & 50.1  \\
\end{tabular}
\caption{The comparison on image understanding ability between LLMs and VLMs, reflected by the average accuracy (\%) on \QAbench{}. We feed the rasterialized vector graphics in PNG format along with the same question in text format to LLaVA-1.5-13b to evaluate its performance on \QAbench{}}
\label{vlms}
\end{table}

\paragraph{Impact of Vector Graphics Sequence Length}
We next study the influence of the length of the vector graphics on vector graphics understanding. Results for GPT-4 are shown in Table~\ref{tab:QABench length}, where GPT-4 shows consistent performance across different length groups. Specifically, low-level vector graphics format such as SVG is most sensitive to the length. When the length increases, the understanding performance on SVG decreases steadily, while the understanding performance on other high level format remains stable. Another noticeable finding is that questions requiring complex reasoning, such as Usage in SVG or Relation in Graphviz, suffer more from the increasing sequence length.

\paragraph{Can LLMs Reason over Vector Graphics?}
The reasoning process of GPT-4 under the CoT setting is shown in Figure~\ref{fig:prompt example}. Results show that GPT-4 can detect the key information over those samples, such as \textit{"two large, similar shapes that could represent the eyepieces ..."}, for correct reasoning. We include the full conversation in Appendix~\ref{sec: detail reasoning examples}.

\section{Conclusion}

Our study unveils new insights into the capabilities of LLMs in understanding and generating vector graphics. We discovered that LLMs demonstrate decent vector graphics understanding in TikZ, Graphviz, and SVGs, with a particular strength in understanding vector graphics code with higher-level semantics. We also found that LLMs often exhibit strong vector graphics generation capabilities. Interestingly, advanced prompting techniques can significantly improve performance for low-level formats such as SVG, and while GPT-4 had the strongest performance, open-source models like Llama-3-70B and Qwen2-72B show competitive performance. Our work lays a groundwork for future studies into LLMs' vector graphics understanding and generation benchmarking, and we hope it will inspire further efforts to enhance these capabilities. We will release our benchmark dataset and evaluation pipeline.

\section{Limitations}
\label{limitations}

We acknowledge that one cannot systematically evaluate the behavior of the closed-source models we employed, namely GPT-4, GPT-35-Turbo, and GPT-4V. Besides, more evaluations on recent LLMs can be conducted, which can provide more supporting experiments on LLMs' behavior on vector graphics understanding and generation. 

Furthermore, recent works propose more prompting techniques such as Tree of Thoughts (ToT)~\cite{yao2024tree} and Everything of Thoughts (XoT)~\cite{ding2023everything}.
Incorporating these prompting techniques could further enhance our study.

\section*{Acknowledgements}

This work was supported in part by NSF CAREER IIS2150012, Adobe Data Science award, Microsoft Accelerate Foundation Models Research Program, and Institute of Information \& communications Technology Planning \& Evaluation (IITP) grants funded by the Korea government (MSIT) (No. 2022-0-00871, Development of AI Autonomy and Knowledge Enhancement for AI Agent Collaboration) and (No. RS-2022-00187238, Development of Large Korean Language Model Technology for Efficient Pre-training).

\bibliography{reference}

\section{Appendix}

\begin{figure*}[htbp]
    \centering
    \includegraphics[width=\linewidth]{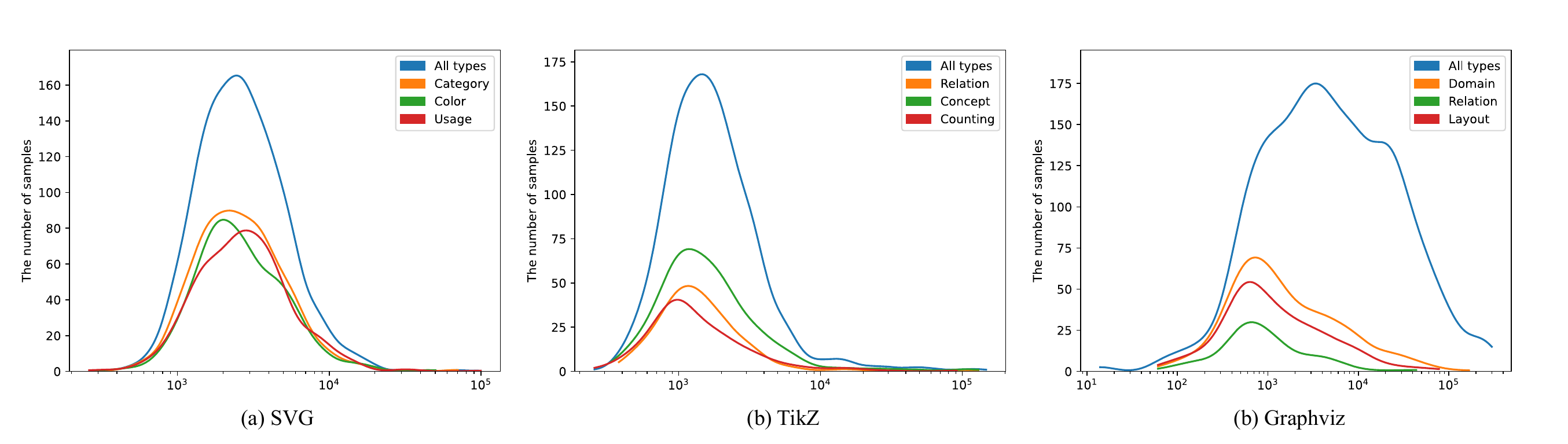}
    \caption{The distribution of vector graphics length in \shortname{}. X-axis denotes the string length of vector graphics files in each vector grapics format.  }
    \label{fig:distribution}
\end{figure*}

\begin{figure*}[!ht]
\centering
\includegraphics[width=\linewidth]{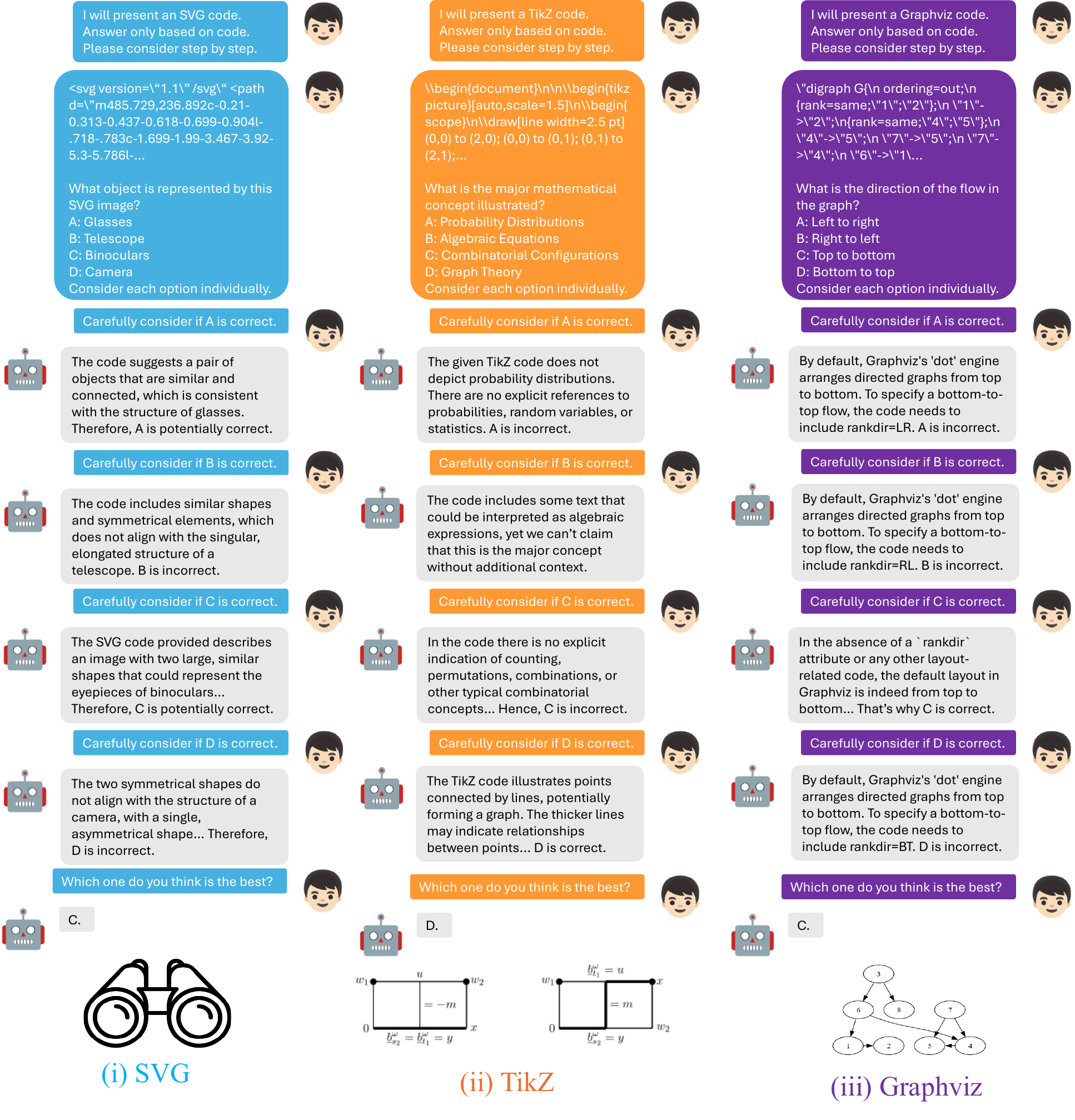}
\caption{We include the full conversation with GPT-4 as indicated in Figure~\ref{fig:prompt example}. We ask the model to consider if each option is correct individually, then ask another GPT-4 model to judge if the reasoning matches the correct answer.}
\label{fig:prompt example full}
\end{figure*}

\subsection{The specific prompt we used}

\subsubsection{Prompts used to build the dataset}

\paragraph{Question Generation}

System prompt: The system prompts used to generate questions are different for different types of vector graphics and different types of questions. See the code in supplemental material for details.

User prompt: \textit{The caption of this image is \{caption\}, generate the json according to the instruction. <IMAGE>}

\paragraph{Caption Generation}

System prompt: \textit{Generate a detailed caption for the given image. The reader of your caption should be able to replicate this picture.}

User prompt: \textit{<IMAGE>}

\subsubsection{Prompts used to evaluate the models's understanding ability}

\paragraph{Zero-shot}

System prompt: \textit{I will present a \{format\} code. Please answer my questions only based on code. Answer and only answer the letter corresponding to the correct option. Do not add any additional comment in your response}

User prompt: \textit{"\{code\}". Given this image, answer \{question\}. Options are \{options\}}

\paragraph{Few-shot}

System prompt: \textit{I will present a \{format\} code. Please answer my questions only based on code. Answer and only answer the letter corresponding to the correct option. Do not add any additional comment in your response. For your reference, I will give you some examples.}

User prompt: \textit{This is an example, the code is: \{code\}}

User prompt: \textit{Given this image, answer \{few\_shot\_sample\_question\}. Options are \{few\_shot\_sample\_options\}}

Simulated assistant prompt: \textit{\{few\_shot\_sample\_answer\}}

Repeat the last three prompts for three times, each time pass a different samples.

User prompt: \textit{"\{code\}". Given this image, answer \{question\}. Options are \{options\}}

\paragraph{Zero-shot-cot}

System prompt: \textit{I will present a \{format\} code. Please answer my questions only based on code. Please consider the question step by step.}

User prompt: \textit{\{code\}}

User prompt: \textit{Given this image, the question is \{question\}. Options are \{options\}. Do not answer directly, consider each option individually.}

User prompt: \textit{Carefully consider if the option A is correct}

Wait for the large language model's reponse and add its response to the context.

User prompt: \textit{Carefully consider if the option B is correct}

Wait for the large language model's reponse and add its response to the context.

User prompt: \textit{Carefully consider if the option C is correct}

Wait for the large language model's reponse and add its response to the context.

User prompt: \textit{Carefully consider if the option D is correct}

Wait for the large language model's reponse and add its response to the context.

User prompt: \textit{Which option is the best? Answer and only answer the letter corresponding to the correct option. Do not add any additional comment in your response}

\subsubsection{Prompts used to evaluate models' generation ability}

System prompt: \textit{Generate a \{format\} based on the caption below. You should output the compilable code without any additional information.}

User prompt: \textit{\{caption\}}

\subsection{Data distribution}
We include the distribution of \QAbench{} grouped by each vector graphic category in Figure~\ref{fig:distribution}, each in itself grouped by the specific question categories we assigned.

\subsection{Detailed examples for reasoning}
\label{sec: detail reasoning examples}
We include the full version of the three example conversations previously put in Figure~\ref{fig:prompt example} now in Figure~\ref{fig:prompt example full}. The three conversations show how we only input the vector graphics code, exhibit the question, ask the model to consider each question carefully, and finally make its best choice.

\subsection{Llama variants used in this paper}

We evaluated Llama's variants, Llama-3-8B-Instruct-262k\footnote{https://huggingface.co/gradientai/Llama-3-8B-Instruct-262k} and  Llama-3-70B-Instruct-Gradient-262k\footnote{https://huggingface.co/gradientai/Llama-3-70B-Instruct-Gradient-262k} in this paper because they have extended context length.

\subsection{Human filtering}

The authors of this study, proficient in English with extensive research experience in vision-language
learning, perform the vector graphics QA filtering.

\subsection{Programs and Data Release}

Our code and data is included in the supplementary materials.

\end{document}